\documentclass[runningheads]{llncs}
\usepackage[T1]{fontenc}
\usepackage{graphicx}
\usepackage{url}

\usepackage{enumitem}
\usepackage{amssymb}
\usepackage{float}
\usepackage{booktabs}
\usepackage{multirow}
\usepackage{siunitx}
\usepackage[table]{xcolor}
\usepackage{threeparttable}
\usepackage{makecell}
\usepackage{tabularx}
\usepackage{array}
\usepackage[protrusion=false]{microtype}
\definecolor{lightyellow}{RGB}{255, 253, 235}
\definecolor{lightblue}{RGB}{230, 240, 255}

\def\BibTeX{{\rm B\kern-.05em{\sc i\kern-.025em b}\kern-.08em
    T\kern-.1667em\lower.7ex\hbox{E}\kern-.125emX}}

\usepackage[most]{tcolorbox}
\usepackage{fontawesome5}
\definecolor{systemColor}{RGB}{230, 240, 255}
\definecolor{inputColor}{RGB}{255, 253, 235}
\definecolor{borderColor}{RGB}{200, 200, 200}

\tcbset{
    promptbox/.style={
        enhanced,
        boxrule=0.5pt,
        colframe=borderColor,
        left=4pt, right=4pt, top=10pt, bottom=4pt,
        fonttitle=\bfseries\small\sffamily,
        coltitle=black,
        attach boxed title to top left={yshift=-10pt, xshift=10pt},
        boxed title style={colback=white, frame hidden},
        breakable,
        arc=2pt
    }
}

\newtcolorbox{systembox}[1][]{promptbox, colback=systemColor, title=\faCog\ System Prompt, #1}
\newtcolorbox{inputbox}[1][]{promptbox, colback=inputColor, title=\faUser\ Input, #1}
    
\begin{document}
\title{How Auxiliary Reasoning Unleashes GUI Grounding in VLMs}
\titlerunning{Auxiliary Reasoning for GUI Grounding}
%
\author{Weiming Li\inst{1}\orcidID{0009-0008-0509-601X} \and
Yan Shao\inst{2}\orcidID{0009-0001-1130-9201} \and
Jing Yang\inst{1}\orcidID{0009-0007-9035-0766} \and
Yujing Lu\inst{1}\orcidID{0009-0005-2941-7357} \and
Ling Zhong\inst{1}\orcidID{0000-0003-0053-2456} \and
Yuhan Wang\inst{1}\orcidID{0009-0005-9080-2764} \and
Min Yu\inst{3}\orcidID{0000-0001-9757-3021} \and
Tongxiao Ruan\inst{2}\orcidID{0009-0002-3320-2971} \and
Manni Duan\inst{1}\orcidID{0009-0003-0333-3994}}
\authorrunning{W. Li et al.}
%
\institute{
Zhejiang Lab, Hangzhou, China \\
\email{\{silang1225, duanmanni\}@gmail.com, \{yangjing0128, luyujing, zhongling, wangyuhan\}@zhejianglab.org}
\and
Hangzhou Research and Development Center, China Mobile, Hangzhou, China \\
\email{\{shaoyan, ruantongxiao\}@cmhi.chinamobile.com}
\and
Innovation Center of Yangtze River Delta, Zhejiang University, Jiaxing, China \\
\email{yumin0307@zju.edu.cn}
}
\maketitle              
\begin{abstract}
Graphical user interface (GUI) grounding is a fundamental task for building GUI agents. However, general vision-language models (VLMs) struggle with this task due to a lack of specific optimization. We identify a key gap in this paper: while VLMs exhibit significant latent grounding potential, as demonstrated by their performance measured by Pointing Game, they underperform when tasked with outputting explicit coordinates. To address this discrepancy and bypass the high data and annotation costs of current fine-tuning approaches, we propose three zero-shot auxiliary reasoning methods. By providing explicit spatial cues such as axes, grids and labeled intersections as part of the input image, these methods enable VLMs to better articulate their implicit spatial understanding capabilities. We evaluate these methods on four GUI grounding benchmarks across seven open-source and proprietary VLMs. Experimental results show substantial gains from auxiliary reasoning. Mark-Grid Scaffold boosts Gemini-3.1-Pro from 11.72\% under direct inference to 95.20\% on ScreenSpot-v2, achieves state-of-the-art performance on ScreenSpot, and approaches the strongest fine-tuned methods on ScreenSpot-v2 and UI-I2E-Bench. Our code is available at \url{https://github.com/liweim/AuxiliaryReasoning}.
\end{abstract}

\section{Introduction}
\label{sec:intro}

Graphical user interface (GUI) grounding is a specialized form of visual grounding that involves locating specific screen components based on natural language instructions. Beyond identification, this task requires the model to output precise screen coordinates, such as a click point or a bounding box. This capability is essential for GUI agents to translate user commands into adequate screen interactions.

Although vision-language models (VLMs) have achieved substantial progress in general multimodal understanding \cite{achiam2023gpt}, they often struggle with GUI grounding \cite{cheng2024seeclick,wu2024atlas,li2025screenspotpro}. Unlike natural images, GUI screenshots contain dense text, fine-grained icons and highly structured layouts. These features are not emphasized during standard VLM pretraining, resulting in poor grounding performance \cite{cheng2024seeclick}. Using \textit{Pointing Game} \cite{zhang2018top}, which evaluates whether the point of highest attention falls within the ground-truth region, we observe that VLMs exhibit considerable latent grounding potential under this diagnostic protocol. However, when the same models are required to output explicit coordinates, their performance drops drastically, revealing a clear gap between latent spatial localization and explicit coordinate prediction. As shown in Figure~\ref{fig:1}(a), the VLM fails to produce accurate coordinates through direct inference, which we refer to as Direct Prediction, whereas Pointing Game correctly identifies the target location via attention peak localization (Figure~\ref{fig:1}(b)). 

Existing efforts have made substantial progress in GUI grounding, but they are typically driven by task-specific supervision, specialized architectures, and large-scale GUI data collection. Early representative systems such as CogAgent \cite{hong2024cogagent} and SeeClick \cite{cheng2024seeclick} demonstrate that GUI grounding can be substantially improved through dedicated training on GUI-centric data. Subsequent work further scales this paradigm, where UGround trains on 1.3 million screenshots with 10 million element references \cite{gou2024navigating}, while AGUVIS scales it to 4.2 million sampled interface elements \cite{xu2024aguvis}. Other representative systems, including GTA1 \cite{yang2026gta}, UI-Ins \cite{chen2025ui}, and UI-Venus \cite{gu2025ui}, further improve GUI grounding through reinforcement learning, test-time scaling, instruction-as-reasoning, and GUI-specific fine-tuning. More recent methods, such as GUI-Actor \cite{wu2025gui}, InfiGUI-G1 \cite{liu2025infigui}, and IRIS \cite{ge2024iris}, improve GUI agents through coordinate-free action prediction, adaptive exploration policies, and specialized reasoning strategies. Although these methods substantially advance GUI grounding, they typically remain tied to costly annotation, model training, or parameter access, which limits their applicability to proprietary foundation models and off-the-shelf VLMs.

In contrast, we explore a training-free paradigm based on \textit{auxiliary reasoning}, where additional spatial cues are introduced at inference time to guide more accurate coordinate prediction without modifying model parameters. This direction is related to training-free methods that introduce explicit spatial cues in general visual grounding. For example, Grid-Augmented Vision \cite{chae2024grid} overlays regular grids on images to provide explicit spatial references, while Scaffold Prompting \cite{lei2024scaffolding} introduces indexed visual anchors to support spatial reasoning in VLMs. These studies suggest that failures in explicit localization may arise not only from insufficient visual understanding, but also from a mismatch between latent spatial knowledge and the form of the required output. Several recent GUI-oriented methods also incorporate stronger spatial guidance, such as attention-driven grounding \cite{xu2025attention}, iterative narrowing mechanisms \cite{nguyen2024improved}, and data-efficient reasoning frameworks \cite{lee2025reguide}. However, these approaches typically rely on more specialized reasoning procedures tailored to the GUI domain. In contrast, our work studies a lightweight zero-shot spatial scaffolding paradigm for GUI grounding that applies to both open-source and proprietary VLMs. Rather than injecting new grounding capability through supervision, this paradigm aims to better elicit and structure the model's existing spatial knowledge for explicit localization.

Following this paradigm, we introduce three lightweight methods, namely \textit{Coordinate Scaffold}, \textit{Axis-Grid Scaffold} and \textit{Mark-Grid Scaffold}, that enhance both open-source and proprietary VLMs with explicit spatial cues. The first two methods leverage visual overlays such as axes, grids and labeled intersections to preserve the model's native reasoning process, while the third simplifies continuous coordinate prediction as discrete grid ID prediction. We evaluate these methods on four GUI grounding benchmarks with seven backbone VLMs. Our methods generally improve click accuracy on seven VLMs across four benchmarks. For instance, Mark-Grid Scaffold elevates the score of Gemini-3.1-Pro from a direct inference score of 11.72\% to 95.20\% on ScreenSpot-v2 and achieves state-of-the-art performance on ScreenSpot. Our contributions are summarized as follows:
\begin{itemize}
    \item A significant performance gap between Direct Prediction and Pointing Game is identified in general VLMs, suggesting substantial latent grounding potential beyond direct coordinate prediction.
    \item Three lightweight zero-shot auxiliary reasoning methods are proposed to enhance GUI grounding in VLMs, with compatibility across both open-source and proprietary models.
    \item Extensive experiments on seven open-source and proprietary VLMs across four GUI benchmarks show substantial gains from auxiliary reasoning. Mark-Grid Scaffold achieves state-of-the-art performance on ScreenSpot, and approaches the strongest fine-tuned methods on ScreenSpot-v2 and UI-I2E-Bench.
\end{itemize}

\section{Unleashing the GUI Grounding Capability}
\label{sec:method}
In this section, we first describe the implementation of Pointing Game, which we use as a diagnostic tool to probe latent spatial grounding potential in VLMs. We then introduce three auxiliary reasoning methods that provide explicit spatial cues at inference time to narrow the gap between latent spatial localization and explicit coordinate prediction.

\subsection{Implementation of Pointing Game}
Our implementation of Pointing Game involves extracting and processing the attention maps from the model's last text token to all image tokens, and then determining whether the corresponding point of highest attention lies within the ground-truth region. Formally, our procedure is defined as follows.

Consider the input prompt: ``Where should I click if I want to \texttt{\{instruction\}}?''. Let $A \in \mathbb{R}^{L \times H \times N \times N}$ be the complete attention tensor, where $L$ is the number of layers, $H$ is the number of attention heads, and $N$ is the total number of tokens (including text, image and special tokens). Let $I \subset \{1, \ldots, N\}$ be the set of indices corresponding to the image tokens, and let $t^{\ast} \in \{1, \ldots, N\}$ be the last text token of the key phrase ``\texttt{\{instruction\}}'' from the input prompt. Let $GT \in \{0,1\}^{H_o \times W_o}$ be the binary ground-truth region, where $H_o$ and $W_o$ are the height and width of the original image. The function $R: \mathbb{R}^{H_i \times W_i} \to \mathbb{R}^{H_o \times W_o}$ is a spatial resizing function.

For each layer $l \in \{1,2,\ldots,L\}$, the following steps are performed:

First, the attention weights from the last text token to all image tokens are extracted and averaged as follows:
\begin{equation}
\bar{a}_l = \frac{1}{H} \sum_{h=1}^H A[l, h, t^{\ast}, I]
\end{equation}
Here, $A[l, h, t^{\ast}, I]$ represents the slice of the attention tensor for layer $l$ and head $h$, where the query token is the last text token at index $t^{\ast}$ and the key tokens are all the image tokens at indices in the set $I$.

Second, the averaged attention vector is reshaped into a spatial grid and then resized to match the original image dimensions:
\begin{equation}
M_l = R(\mathrm{Reshape}_{H_i \times W_i}(\bar{a}_l))
\end{equation}

Third, the coordinate of the maximum value in the processed attention map is extracted:
\begin{equation}
(x_l, y_l) = \arg\max_{(x,y)} M_l(x,y)
\end{equation}

Finally, it is determined whether the predicted coordinate falls within the ground-truth region:
\begin{equation}
I_l = \mathbb{1}\{GT(x_l,y_l) = 1\}
\end{equation}
where $\mathbb{1}$ is the indicator function that equals 1 if the condition is satisfied and 0 otherwise.

Based on the layer-wise indicators $\{I_l\}_{l=1}^L$, we define three Pointing Game variants:
\begin{equation}
P_{\text{union}} = \bigvee_{l=1}^{L} I_l
\end{equation}
\begin{equation}
P_{\text{mean}} = \frac{1}{L}\sum_{l=1}^{L} I_l
\end{equation}
\begin{equation}
P_{\text{layer}} = I_{l^*}
\end{equation}
where $P_{\text{union}}$ regards a prediction as correct if at least one layer attends to a point inside the ground-truth region, $P_{\text{mean}}$ measures the average success rate across layers, and $P_{\text{layer}}$ uses one fixed layer $l^*$ for each model, chosen empirically for that model.

This implementation is best viewed as a diagnostic protocol rather than a realistic estimate of deployable grounding performance. Taken together, these three variants provide complementary evidence about latent grounding ability by showing whether grounding signals emerge in any layer, can be recovered from a representative fixed layer, or are distributed more consistently across layers.

\subsection{Auxiliary Reasoning Methods}

We propose three auxiliary reasoning methods to enhance the performance of VLMs in GUI grounding: Coordinate Scaffold, Axis-Grid Scaffold and Mark-Grid Scaffold. These methods provide additional spatial cues during inference, which improves the model's ability to accurately locate and reason about objects within images.

\begin{figure*}[!ht]
\centering
\includegraphics[width=\linewidth]{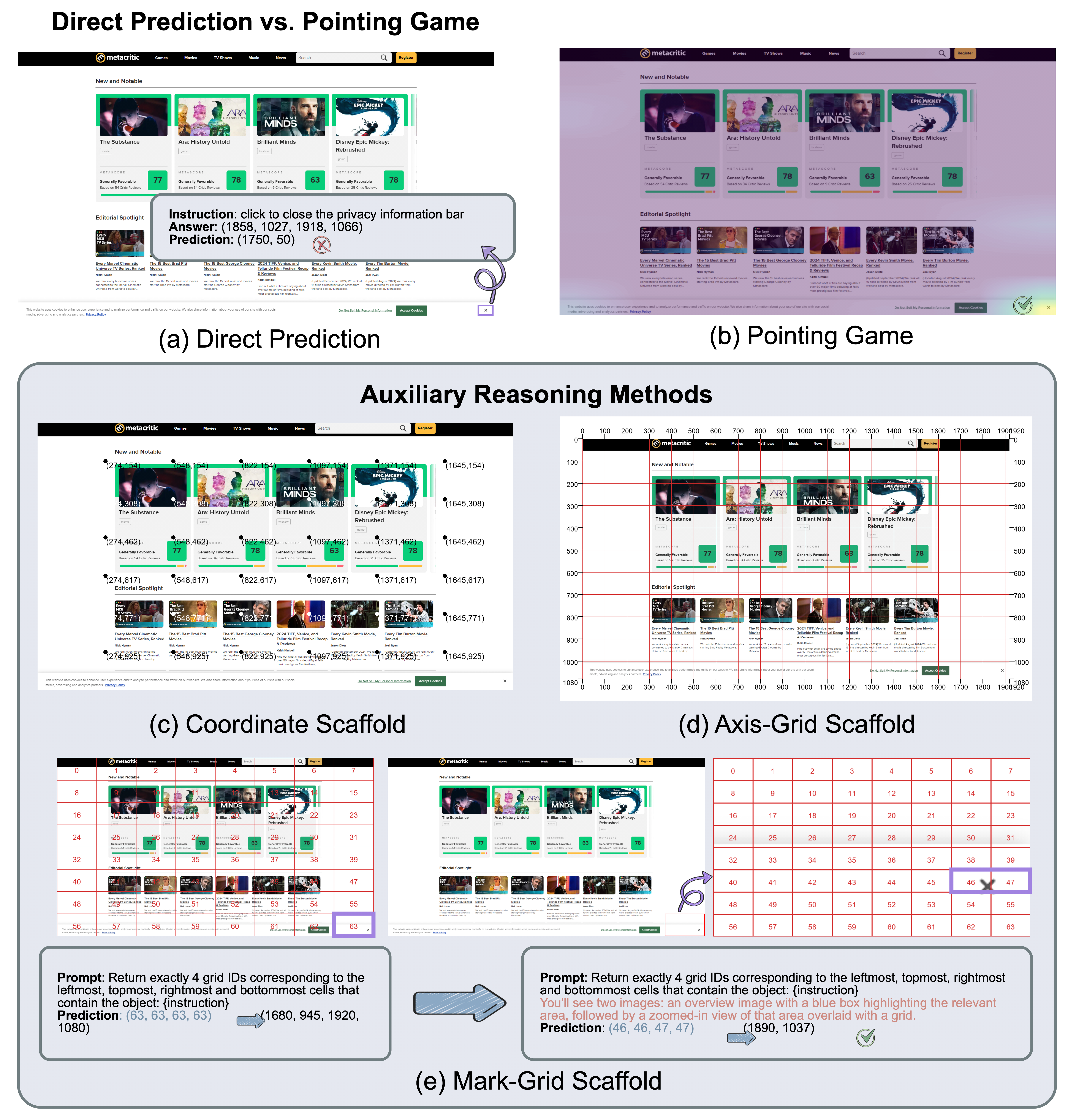}
\caption{This figure compares different GUI grounding methods on a screenshot sample. (a) Direct Prediction shows a model's attempt to directly output coordinates. (b) Pointing Game illustrates a model's latent grounding ability by identifying the point of highest attention. (c) Coordinate Scaffold, (d) Axis-Grid, and (e) Mark-Grid Scaffold are the proposed auxiliary reasoning methods.}
\label{fig:1}
\end{figure*}

\subsubsection{Coordinate Scaffold}

Coordinate Scaffold builds upon the existing Scaffold Prompting method \cite{lei2024scaffolding}. While the original Scaffold Prompting method labels the anchor points with row and column indices, our modification uses the actual corresponding $x,y$ coordinates of the anchor points within the image. This approach directly aligns with predicting explicit coordinates for GUI grounding. Figure~\ref{fig:1}(c) illustrates an example of this method.

The prompt consists of a system message and a user prompt, shown below.

\begin{systembox}[title=System Prompt]
The image is overlaid with an $8\times8$ dot matrix to help you complete the task.\\
1. When referring to any key object in the image, first identify its nearest dot(s), then describe the object.\\
2. Use the dots to determine spatial relationships among objects.\\
3. You may search and reason region by region with the help of the dots.
\end{systembox}

\begin{inputbox}[title=User Prompt]
In the image, where should I click if I want to \{instruction\}?\\
The output coordinate should be in the format: (x, y), for example: (100, 100).\\
The width and height of the image are \{width\} and \{height\}. The origin is the top-left corner.
\end{inputbox}

\subsubsection{Axis-Grid Scaffold}

Axis-Grid Scaffold enhances a model's spatial perception by adding a structured grid overlay. It places coordinate scales at 100-unit intervals on the image's four edges and a corresponding grid across the image. This combination provides a detailed, explicit spatial structure that facilitates the model's reasoning about object locations and relationships. An example is shown in Figure~\ref{fig:1}(d).

The prompt used is shown below.

\begin{inputbox}[title=User Prompt]
In the image, where should I click if I want to \{instruction\}?\\
The output coordinate should be in the format: (x, y), for example: (100, 100).\\
The axis shown in the image is provided to help determine precise positions.\\
Note: a red coordinate grid overlay has been added to assist in precise localization.
\end{inputbox}

\subsubsection{Mark-Grid Scaffold}

Marked-Grid Scaffold simplifies continuous coordinate prediction as a discrete grid ID prediction task. This is achieved by overlaying an $8\times8$ grid on the input image, where each grid cell is labeled with a unique ID at its center. During inference, the model identifies the grid IDs corresponding to the object's four extremities: leftmost, topmost, rightmost and bottommost. This initial bounding box is then used to crop and magnify the region of interest. The cropped region is proportionally resized to 512 pixels on its shorter side and the same $8\times8$ grid is applied. The model is then presented with both the original image annotated with the predicted bounding box and the magnified, grid-marked crop. The model again predicts the four grid IDs defining the object's extremities within this finer-grained view. The final object coordinates are determined by calculating the center coordinates from these four grid IDs. An example of this process is shown in Figure~\ref{fig:1}(e).

The prompts used in the two rounds are given below.

\begin{inputbox}[title=Round-1 User Prompt]
A UI image is overlaid with a red-labeled grid, where each cell is marked with its ID at the center. Determine which grid cell(s) contain the UI element needed to complete the following instruction: \{instruction\}.\\

List exactly 4 grid IDs corresponding to the leftmost, topmost, rightmost, and bottommost cells that contain the target element. These IDs may be identical if the element lies entirely within a single cell. The grid IDs should be between 0 and 63.\\

The output should be in the format: [id1, id2, id3, id4], for example: [3, 3, 4, 9].
\end{inputbox}

\begin{inputbox}[title=Round-2 User Prompt]
A UI image is overlaid with a red-labeled grid, where each cell is marked with its ID at the center. Determine which grid cell(s) contain the UI element needed to complete the following instruction: \{instruction\}.\\

List exactly 4 grid IDs corresponding to the leftmost, topmost, rightmost, and bottommost cells that contain the target element. These IDs may be identical if the element lies entirely within a single cell. The grid IDs should be between 0 and 63.\\

The output should be in the format: [id1, id2, id3, id4], for example: [3, 3, 4, 9].\\

You will see two images: first, an overview image with a red box highlighting the relevant region; second, a zoomed-in view of that region overlaid with the same labeled grid.
\end{inputbox}

\section{Experiments}
\label{sec:experiments}
\subsection{Experimental Settings}
This section describes the experimental setup used to evaluate the performance of our proposed auxiliary reasoning methods. 

\subsubsection{Benchmarks}
We evaluate the methods on four GUI grounding benchmarks: 
\begin{itemize}
\item \textbf{ScreenSpot} \cite{cheng2024seeclick}: A benchmark for GUI visual grounding, containing 1,272 single-step instructions with corresponding target elements. It covers a range of GUI platforms, including mobile, desktop and web environments. 
\item \textbf{ScreenSpot-v2} \cite{wu2024atlas}: A corrected version of ScreenSpot, which addresses annotation errors and ambiguous instructions while maintaining the same number of samples.
\item \textbf{ScreenSpot-Pro} \cite{li2025screenspotpro}: A challenging benchmark for professional scenarios, featuring 1,581 expert-annotated tasks across 23 applications and three operating systems. This benchmark utilizes ultra-high-resolution screenshots that often exceed 4K dimensions and significantly surpass the standard input resolution limits of vision encoders, which necessitates aggressive downsampling that results in severe visual information loss.
\item \textbf{UI-I2E-Bench} \cite{liu2025ui}: It evaluates GUI grounding through natural language instructions aligned with diverse UI elements. This dataset comprises 1,477 instruction-element pairs and evaluates a model's capability to locate diverse elements within complex contexts.
\end{itemize}

\subsubsection{Metrics}
We use click accuracy as the evaluation metric. Given $N$ test samples, let $(x_i, y_i)$ denote the predicted click point for sample $i$, and let $B_i$ denote the ground-truth bounding box of the target GUI element. Click accuracy is defined as
\[
\text{Acc}=\frac{1}{N}\sum_{i=1}^{N}\mathbf{1}\big[(x_i,y_i)\in B_i\big],
\]
where $\mathbf{1}[\cdot]$ equals 1 if the predicted point falls inside the target bounding box and 0 otherwise. This follows the standard evaluation protocol used in prior GUI grounding work~\cite{li2022grounded}.

\subsubsection{Baselines}
In addition to the original Scaffold Prompting, two baselines are employed to compare with the proposed approaches:

\begin{itemize}
\item \textbf{Direct Prediction}: This baseline uses the original images without any modifications, providing a reference point for model performance.
\item \textbf{Grid-Augmented Vision} \cite{chae2024grid}: This method overlays a $9\times9$ black grid pattern onto the input images. The grid provides explicit spatial references to enhance the model's spatial localization capabilities.
\item \textbf{Scaffold Prompting} \cite{lei2024scaffolding}: This method overlays a matrix of anchor points labeled by row and column indices, encouraging the model to reason about target locations through symbolic spatial references.
\end{itemize}

\subsection{Results}
\subsubsection{Pointing Game versus Direct Prediction}
In this experiment, we compare Direct Prediction with three Pointing Game variants, namely PG-Mean, PG-Layer, and PG-Union, across four benchmarks. As shown in Table~\ref{tab:1}, the Pointing Game variants generally outperform Direct Prediction on open-source models, especially PG-Union and PG-Layer. Among them, PG-Union performs best overall, followed by PG-Layer. These results suggest that the evaluated VLMs already encode useful grounding evidence, but often struggle to translate it into explicit click coordinates. The consistent ordering of PG-Union, PG-Layer, and PG-Mean further indicates that such evidence is unevenly distributed across layers, rather than being stably available throughout the network.

\begin{table}[!p]
\centering
\caption{Main results on four GUI grounding benchmarks. All values are reported as click accuracy (\%). Our proposed methods are highlighted in gray. The best results are in bold, and the second-best are underlined. Dashes (--) denote unavailable results. For open-source VLMs, we additionally report three Pointing Game variants, namely PG-Mean, PG-Layer, and PG-Union. PG-Layer uses fixed layers 18, 12, and 4 for Gemma-3-12B, Gemma-3-4B, and SigLip-400M, respectively. As a CLIP-style model, SigLip is evaluated only with Pointing Game variants.}
\label{tab:1}
\resizebox{\textwidth}{!}{

\begin{threeparttable}
\begin{tabular}{l *{7}{r}}
\toprule
\multicolumn{1}{c}{\textbf{Method}} & 
\multicolumn{4}{c}{\textbf{Proprietary}} & 
\multicolumn{3}{c}{\textbf{Open-Source}} \\
\cmidrule(lr){2-5} \cmidrule(lr){6-8}
 & {\makecell{\textbf{Gemini-}\\\textbf{3.1-Pro} }} & {\makecell{\textbf{Gemini-}\\\textbf{2.5-Flash} }} & {\makecell{\textbf{GPT-4o}}} & {\makecell{\textbf{Claude-}\\\textbf{3.5-Sonnet}}} & {\makecell{\textbf{Gemma-}\\\textbf{3-12B}}} & {\makecell{\textbf{Gemma-}\\\textbf{3-4B}}} & {\makecell{\textbf{SigLip-}\\\textbf{400M}}} \\
\midrule

\rowcolor{lightblue}
\multicolumn{8}{l}{\textbf{ScreenSpot}} \\

\midrule
Direct Prediction & 15.62 & 5.50 & 20.83 & 23.98 & 9.59 & 4.95 & {--} \\

PG-Mean & {--} & {--} & {--} & {--} & 9.30 &	10.13 &	1.58  \\

PG-Layer & {--} & {--} & {--} & {--} & 29.01 & 28.54 & 2.12 \\

PG-Union & {--} & {--} & {--} & {--} & \textbf{47.56} & \textbf{46.78} & \textbf{24.92} \\
\rowcolor[gray]{0.95}
Coordinate Scaffold & \underline{22.66} & 34.04 & 24.61 & 34.51 & 16.04 & 11.32 & {--} \\
\rowcolor[gray]{0.95}
Axis-Grid Scaffold & 20.31 & \underline{56.92} & \underline{45.75} & \underline{49.76} & 26.34 & 13.21 & {--} \\
\rowcolor[gray]{0.95}
Mark-Grid Scaffold & \textbf{92.85} & \textbf{69.10} & \textbf{62.03} & \textbf{65.25} & \underline{32.00} & \underline{22.09} & {--} \\
\midrule

\rowcolor{lightblue}
\multicolumn{8}{l}{\textbf{ScreenSpot-v2}} \\

\midrule
Direct Prediction & 11.72 & 5.66 & 20.83 & 22.80 & 7.94 & 4.40 & {--} \\

PG-Mean & {--} & {--} & {--} & {--} & 8.62 &	9.41 &	1.50   \\

PG-Layer & {--} & {--} & {--} & {--} & 27.59 & 26.57 & 2.04 \\

PG-Union & {--} & {--} & {--} & {--} & \textbf{46.23} & \textbf{45.52} & \textbf{24.21} \\
\rowcolor[gray]{0.95}
Coordinate Scaffold & 18.75 & 35.30 & 24.61 & 34.51 & 15.64 & 11.08 & {--} \\
\rowcolor[gray]{0.95}
Axis-Grid Scaffold & \underline{21.88} & \underline{56.37} & \underline{47.56} & \underline{50.24} & 25.79 & 13.84 & {--} \\
\rowcolor[gray]{0.95}
Mark-Grid Scaffold & \textbf{95.20} & \textbf{72.09} & \textbf{62.97} & \textbf{67.77} & \underline{33.41} & \underline{22.64} & {--} \\
\midrule

\rowcolor{lightblue}
\multicolumn{8}{l}{\textbf{ScreenSpot-pro}} \\

\midrule
Direct Prediction & 0.95 & 2.02 & 1.27 & 1.20 & 0.06 & 0.13 & {--} \\

PG-Mean & {--} & {--} & {--} & {--} & 0.10 &	0.10 &	0.04   \\

PG-Layer & {--} & {--} & {--} & {--} & 0.44 & 0.32 & 0.13 \\

PG-Union & {--} & {--} & {--} & {--} & \underline{1.71} & \textbf{1.20} & \textbf{1.01} \\
\rowcolor[gray]{0.95}
Coordinate Scaffold & \underline{3.77} & 1.77 & 0.25 & 0.70 & 0.32 & 0.13 & {--} \\
\rowcolor[gray]{0.95}
Axis-Grid Scaffold & 0.63 & \underline{4.36} & \underline{1.96} & \underline{1.83} & 0.25 & 0.06 & {--} \\
\rowcolor[gray]{0.95}
Mark-Grid Scaffold & \textbf{47.80} & \textbf{23.47} & \textbf{12.52} & \textbf{17.33} & \textbf{2.28} & \underline{0.82} & {--} \\
\midrule

\rowcolor{lightblue}
\multicolumn{8}{l}{\textbf{UI-I2E-Bench}} \\

\midrule
Direct Prediction & 10.14 & 5.42 & 15.23 & 16.38 & 4.67 & 3.32 & {--} \\

PG-Mean & {--} & {--} & {--} & {--} & 2.42 &	2.59 &	1.10   \\

PG-Layer & {--} & {--} & {--} & {--} & 8.73 & 7.72 & 1.56 \\

PG-Union & {--} & {--} & {--} & {--} & \textbf{22.07} & \textbf{19.43} & \textbf{17.60} \\
\rowcolor[gray]{0.95}
Coordinate Scaffold & \underline{18.92} & 21.53 & 12.05 & 23.70 & 6.77 & 4.54 & {--} \\
\rowcolor[gray]{0.95}
Axis-Grid Scaffold & 16.22 & \underline{40.49} & \underline{34.94} & \underline{41.03} & 13.00 & 7.11 & {--} \\
\rowcolor[gray]{0.95}
Mark-Grid Scaffold & \textbf{81.76} & \textbf{49.29} & \textbf{41.16} & \textbf{45.63} & \underline{15.71} & \underline{10.83} & {--} \\
\bottomrule
\end{tabular}
\end{threeparttable}
}
\end{table}

\subsubsection{Auxiliary Reasoning Methods}
Table~\ref{tab:1} presents the main comparison of Direct Prediction, three Pointing Game variants, and three proposed auxiliary reasoning methods across four GUI grounding benchmarks. Overall, the auxiliary reasoning methods generally improve over Direct Prediction for all evaluated VLMs. Among them, Mark-Grid Scaffold delivers the largest gains. For example, on ScreenSpot-v2, it raises the performance of Gemini-3.1-Pro from 11.72\% under Direct Prediction to 95.20\%. We attribute this improvement to its progressive coarse-to-fine reasoning loop, which decomposes direct coordinate prediction into a sequence of easier localization steps. More broadly, the three methods form a clear progression in the amount of structural guidance they provide. Coordinate Scaffold supplies explicit coordinate anchors, but still leaves the model to infer global spatial relations in a largely one-shot manner. Axis-Grid Scaffold further regularizes the image into a more interpretable coordinate space, making spatial referencing easier while preserving single-round inference. Mark-Grid Scaffold goes one step further by iteratively narrowing the search region, thereby reducing both the effective search space and the difficulty of precise coordinate prediction.

Another notable trend is that proprietary VLMs benefit more substantially from auxiliary reasoning than open-source models, suggesting that the effectiveness of the proposed scaffolds depends not only on the spatial cues themselves but also on the model's underlying multimodal reasoning ability.

On the more challenging ScreenSpot-Pro benchmark, performance drops substantially for all methods. We attribute this decline primarily to the ultra-high-resolution screenshots in ScreenSpot-Pro, which are inevitably downsampled by the visual encoders of current VLMs and make fine-grained UI elements difficult to distinguish. Despite this limitation, the Mark-Grid Scaffold still achieves substantial improvements, suggesting that structured auxiliary reasoning can partially mitigate the difficulty of coordinate prediction under constrained visual input. We note that Mark-Grid Scaffold outperforms the Pointing Game variants for Gemma-3-12B, likely because its coarse-to-fine zooming can recover local details that are lost in single-pass Pointing Game evaluation.

Table~\ref{tab:specialist_compare} compares our method with representative fine-tuned and training-free GUI grounding methods. Mark-Grid Scaffold achieves state-of-the-art performance on ScreenSpot and obtains the second-best results on ScreenSpot-v2 and UI-I2E-Bench, approaching the strongest fine-tuned GUI grounding methods without task-specific training. These results highlight the effectiveness of structured auxiliary reasoning as a training-free alternative to GUI-specific fine-tuning.

\begin{table}[H]
\centering
\caption{Comparison with representative GUI grounding methods from prior work. All values are reported as click accuracy (\%). All training-free methods are implemented using Gemini-3.1-Pro. Our proposed method is highlighted in gray. The best results are in bold, and the second-best are underlined.}
\label{tab:specialist_compare}
\small
\setlength{\tabcolsep}{5pt}
\renewcommand{\arraystretch}{1.10}
\resizebox{\textwidth}{!}{
\begin{tabular}{lrrrr}
\toprule
\textbf{Method} & \textbf{ScreenSpot} & \textbf{ScreenSpot-v2} & \textbf{ScreenSpot-Pro} & \textbf{UI-I2E-Bench} \\
\midrule

\rowcolor{lightblue}
\multicolumn{5}{l}{\textbf{Fine-tuned}} \\
CogAgent-18B \cite{hong2024cogagent} & 47.4 & 52.8 & 7.7 & -- \\
SeeClick-7B \cite{cheng2024seeclick} & 55.8 & 55.1 & 1.1 & 26.4 \\
OS-ATLAS-4B \cite{wu2024atlas} & 70.1 & 68.5 & 3.7 & 44.3 \\
UGround-7B \cite{gou2024navigating} & 74.1 & 76.3 & 16.5 & 54.2 \\
ShowUI-2B \cite{lin2025showui} & 76.8 & 77.3 & 7.7 & 41.5 \\
OS-ATLAS-7B \cite{wu2024atlas} & 82.5 & 84.1 & 18.9 & 58.6 \\
Aguvis-7B \cite{xu2024aguvis} & 84.4 & 86.0 & 22.9 & 53.2 \\
Jedi-3B \cite{xie2026scaling} & -- & 88.6 & 36.1 & -- \\
InfiGUI-R1-3B \cite{liu2025infigui} & 87.5 & -- & 35.7 & 69.7 \\
UI-TARS-1.5-7B \cite{qin2025ui} & 88.1 & 89.7 & 42.0 & 73.2 \\
UI-TARS-72B \cite{qin2025ui} & 88.4 & 90.3 & 38.1 & 73.7 \\
UI-R1-E-3B \cite{lu2026ui} & 89.2 & 89.5 & 33.5 & 69.1 \\
UI-TARS-7B \cite{qin2025ui} & 89.5 & 91.6 & 35.7 & 61.4 \\
UGround-V1-72B \cite{gou2024navigating} & \underline{89.7} & -- & 34.5 & 76.3 \\
UI-Ins-7B \cite{chen2025ui} & -- & 91.6 & 52.2 & 81.1 \\
Jedi-7B \cite{xie2026scaling} & -- & 91.7 & 39.5 & -- \\
GTA1-7B \cite{yang2026gta} & -- & 92.4 & 50.1 & -- \\
GTA1-32B \cite{yang2026gta} & -- & 93.2 & 53.6 & -- \\
UI-Venus-7B \cite{gu2025ui} & -- & 94.1 & 50.8 & -- \\
GTA1-72B \cite{yang2026gta} & -- & 94.8 & \underline{58.4} & -- \\
UI-Ins-32B \cite{chen2025ui} & -- & 94.9 & 57.0 & \textbf{87.3} \\
UI-Venus-72B \cite{gu2025ui} & -- & \textbf{95.3} & \textbf{61.9} & -- \\
\rowcolor{lightblue}
\multicolumn{5}{l}{\textbf{Training-free}} \\
Grid-Augmented Vision & 11.7 & 10.9 & 1.3 & 9.5 \\
Scaffold Prompting & 40.6 & 50.8 & 11.9 & 25.0 \\
\rowcolor[gray]{0.95}
Mark-Grid Scaffold & \textbf{92.3} & \underline{95.2} & 47.8 & \underline{81.8} \\
\bottomrule
\end{tabular}
}
\end{table}

To evaluate generalization across diverse platforms, Table~\ref{tab:domain_results} provides a domain-wise breakdown of the performance of Gemini-3.1-Pro on ScreenSpot-v2, together with the average inference time per sample. Mark-Grid Scaffold performs best across PC, Mobile, and Web, suggesting that its coarse-to-fine design is not tied to a specific interface style but transfers across different layout structures and element densities. The gains are especially pronounced on Mobile and Web, where interface elements are often smaller, denser, and more tightly arranged, making one-shot coordinate prediction particularly difficult. 

\begin{table}[!ht]
\centering
\caption{Performance of Gemini-3.1-Pro on ScreenSpot-v2 across different domains. The last column reports the average inference time per sample. }
\label{tab:domain_results}
\setlength{\tabcolsep}{6pt}
\renewcommand{\arraystretch}{1.10}
\begin{tabular}{lrrrr}
\toprule
\textbf{Method} & \textbf{PC} & \textbf{Mobile} & \textbf{Web} & \textbf{Avg. Time (s)} \\
\midrule
Direct Prediction & 7.50 & 24.39 & 4.26 & \textbf{7.18} \\
Grid-Augmented Vision & 10.00 & 19.51 & 4.26 & 9.26 \\
Scaffold Prompting & 42.50 & 51.22 & 57.45 & 14.27 \\
\textbf{Mark-Grid Scaffold} & \textbf{96.41} & \textbf{95.81} & \textbf{93.59} & 27.00 \\
\bottomrule
\end{tabular}
\end{table}

\subsubsection{Ablation Studies}
\begin{figure}[!ht]
\centering
\includegraphics[width=\linewidth]{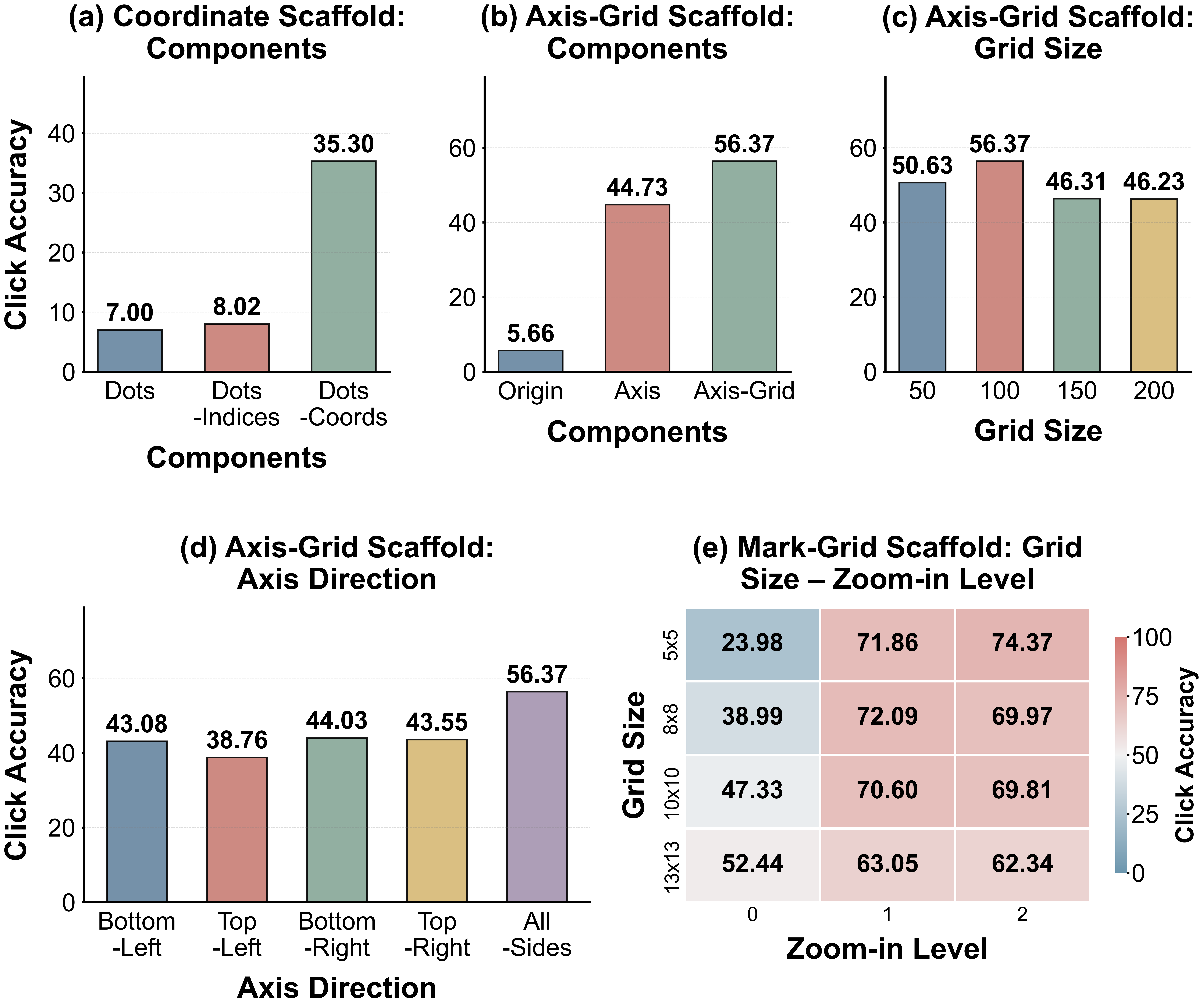}
\caption{Ablation study results on the ScreenSpot-v2 benchmark, using the Gemini-2.5-Flash. The figures show the performance of: (a) Coordinate Scaffold components. (b) Axis-Grid Scaffold components. (c) Impact of Grid Size on Axis-Grid Scaffold. (d) Effect of Axis Direction on Axis-Grid Scaffold. (e) Impact of different groups of Grid Sizes and Zoom-in Levels on Mark-Grid Scaffold.}
\label{fig:3}
\end{figure}

To understand the contribution of each component to the proposed auxiliary reasoning methods, we conduct a series of ablation studies on ScreenSpot-v2 using Gemini-2.5-Flash as a representative backbone. The results are summarized in Figure~\ref{fig:3}.

In Figure~\ref{fig:3}(a), we evaluate three configurations for Coordinate Scaffold. The simplest setup \textit{Dots} places anchor points without any supplementary cues. The standard Scaffold Prompting approach \textit{Dots+Indices} augments these points with symbolic indices. Our proposed method \textit{Dots+Coords} annotates each point with its explicit numerical coordinates. The evaluation results indicate that the explicit coordinate annotation leads to the highest accuracy.

In Figure~\ref{fig:3}(b), we compare three variants of Axis-Grid Scaffold. The \textit{Origin} setup refers to the unmodified original image. The \textit{Axis} configuration extends this by adding coordinate scales at 100-pixel intervals along the four edges. The proposed method \textit{Axis+Grid} further incorporates a full grid structure overlaid on the image. This combination is shown to achieve the best result.

We further investigate the impact of grid size on the Axis-Grid Scaffold performance. As shown in Figure~\ref{fig:3}(c), we test four different grid sizes: 50, 100, 150 and 200. The grid size of 100 achieves the highest score.

Next, we analyze the effect of axis direction by placing the axes on different sides of the image. Figure~\ref{fig:3}(d) compares five configurations: \textit{Bottom-Left}, \textit{Top-Left}, \textit{Bottom-Right}, \textit{Top-Right} and \textit{All Sides}. The \textit{All Sides} configuration performs best.

Finally, for Mark-Grid Scaffold, we conduct experiments with different grid sizes ($5\times5$, $8\times8$,  $10\times10$ and $13\times13$) and varying zoom-in levels (0, 1 and 2). Figure~\ref{fig:3}(e) illustrates an inverse correlation between grid size and the number of zoom-in levels. While a $5\times5$ grid with two zoom-in levels achieves the best overall performance, an $8\times8$ grid with a single zoom-in level offers a better balance of performance and computational efficiency. This configuration is therefore chosen as the default for our method.

The ablation studies demonstrate that effective visual auxiliary cues for GUI grounding are characterized by three key qualities: explicitness, integration, and balanced granularity. Explicit spatial information, such as direct coordinate references, proves to be more effective than symbolic cues. Furthermore, integrating multiple complementary visual elements yields better performance than singular components. Finally, the granularity of visual aids should be carefully chosen to avoid both excessive simplicity and visual clutter.

\section{Conclusion}
This work shows that general VLMs have substantial latent potential for GUI grounding, but direct inference cannot fully realize it. The three Pointing Game variants reveal the gap between latent localization and explicit coordinate prediction. By introducing lightweight spatial cues such as grids, axes, and labeled intersections, our zero-shot auxiliary reasoning methods substantially narrow this gap without fine-tuning. In particular, Mark-Grid Scaffold achieves state-of-the-art performance on ScreenSpot and approaches the strongest fine-tuned methods on ScreenSpot-v2 and UI-I2E-Bench, while requiring no parameter updates. Although effectiveness varies across models and settings, these methods offer a practical way to enhance GUI grounding for both open-source and proprietary VLMs.

\begin{credits}

\subsubsection{\discintname}
The authors declare that they have no known competing financial interests or personal relationships that could have appeared to influence the work reported in this paper.
\end{credits}

\bibliographystyle{splncs04}
\bibliography{refs}

\end{document}